\documentclass[sigconf]{acmart}
\usepackage{multirow}
\usepackage{color}
\usepackage{amsfonts}
\usepackage{enumerate}
\usepackage{setspace}

\newcommand{\stitle}[1]{\vspace*{0em}\noindent{\bf #1:\/}}

\AtBeginDocument{%
  \providecommand\BibTeX{{%
    \normalfont B\kern-0.5em{\scshape i\kern-0.25em b}\kern-0.8em\TeX}}}

\begin{document}

\title{InfoBehavior: Self-supervised Representation Learning for Ultra-long Behavior Sequence via Hierarchical Grouping}

\author{Runshi Liu,~~~~Pengda Qin,~~~~Yuhong Li,~~~~Weigao Wen,~~~~Dong Li,~~~~Kefeng Deng,~~~~Qiang Wu}
\email{{runshi.lrs,pengda.qpd,daniel.lyh,weigao.wwg,shiping,kefeng.deng,qiangwu.wq}@alibaba-inc.com}
\affiliation{%
  \institution{Alibaba Group}
  }

%\author{Runshi Liu~~~~Pengda Qin}
%\email{runshi.lrs@alibaba-inc.com}
%\affiliation{%
%  \institution{Alibaba Group}
%  %\streetaddress{30 Shuangqing Rd}
%  \city{Hangzhou}


\begin{abstract}
E-commerce companies have to face abnormal sellers who sell potentially-risky products. 
Typically, the risk can be identified by jointly considering product content (e.g., title and image) and seller behavior.
This work focuses on the behavior feature extraction as behavior sequences can provide valuable clues for the risk discovery by reflecting the sellers' operation habits.
Traditional feature extraction techniques heavily depend on domain experts and adapt poorly to new tasks.
In this paper, we propose a self-supervised method \textbf{InfoBehavior} to automatically extract meaningful representations from ultra-long raw behavior sequences instead of the costly feature selection procedure. 
InfoBehavior utilizes Bidirectional Transformer as feature encoder due to its excellent capability in modeling long-term dependency. However, it is intractable for commodity GPUs because the time and memory required by Transformer grow quadratically with the increase of sequence length.
Thus, we propose a hierarchical grouping strategy to aggregate ultra-long raw behavior sequences to length-processable high-level embedding sequences.
Moreover, we introduce two types of pretext tasks. 
\textcolor{black}{\emph{Sequence-related}} pretext task defines a contrastive-based training objective to correctly select the masked-out coarse-grained/fine-grained behavior sequences against other "distractor" behavior sequences;
\textcolor{black}{\emph{Domain-related}} pretext task designs a classification training objective to correctly predict the domain-specific statistical results of anomalous behavior.
We show that behavior representations from the pre-trained InfoBehavior can be directly used or integrated with features from other side information to support a wide range of downstream tasks. 
Experimental results demonstrate that InfoBehavior significantly improves the performance of Product Risk Management and Intellectual Property Protection.

\end{abstract}





\maketitle

\section{introduction}
\label{sec:introduction}

	Recently, e-commerce companies have paid increasing attention to the application of artificial intelligence techniques in the area of risk management. To protect consumers' rights and interests, the detection of risky products posted by abnormal sellers is vital for transaction security in the e-commerce scenario.
	Typically, the risk can be identified from the content (e.g., title and image) of the product and the behavior of the seller.
	Practice proves that content-related data suffers from the problem of temporal effectiveness and high-risk confrontation: abnormal sellers can bypass the detectors by repeatedly modifying the form of content expression. 
	Behavior-related data is always accompanied by spatial and temporal attributes.
	Behavior sequences can reflect the latent operation habits of sellers, which is extremely high-cost for sellers to confront. 
	The longer behavior sequences provide more valuable clues for anomaly detection, but also impose higher requirements of feature selection. Traditional supervised methods \citep{olszewski2014fraud,argyriou2014fraud} are over-reliance on the empirical knowledge and hard to model spatiotemporal information; \textcolor{black}{More than that, plenty of efforts are spent on manually analyzing risky behaviors and annotating large-scale labeled data.} Unsupervised learning techniques \citep{chang2007wirevis,chang2008scalable,schaefer2011visual}, such as clustering method, also require substantial efforts to select and build features. \textcolor{black}{Besides, the representations from these methods exist certain limitations in terms of generality.} 
	
	\begin{figure}[t]
		\setlength{\belowcaptionskip}{-0.5cm}
		\centering
		\includegraphics[width=0.9\linewidth]{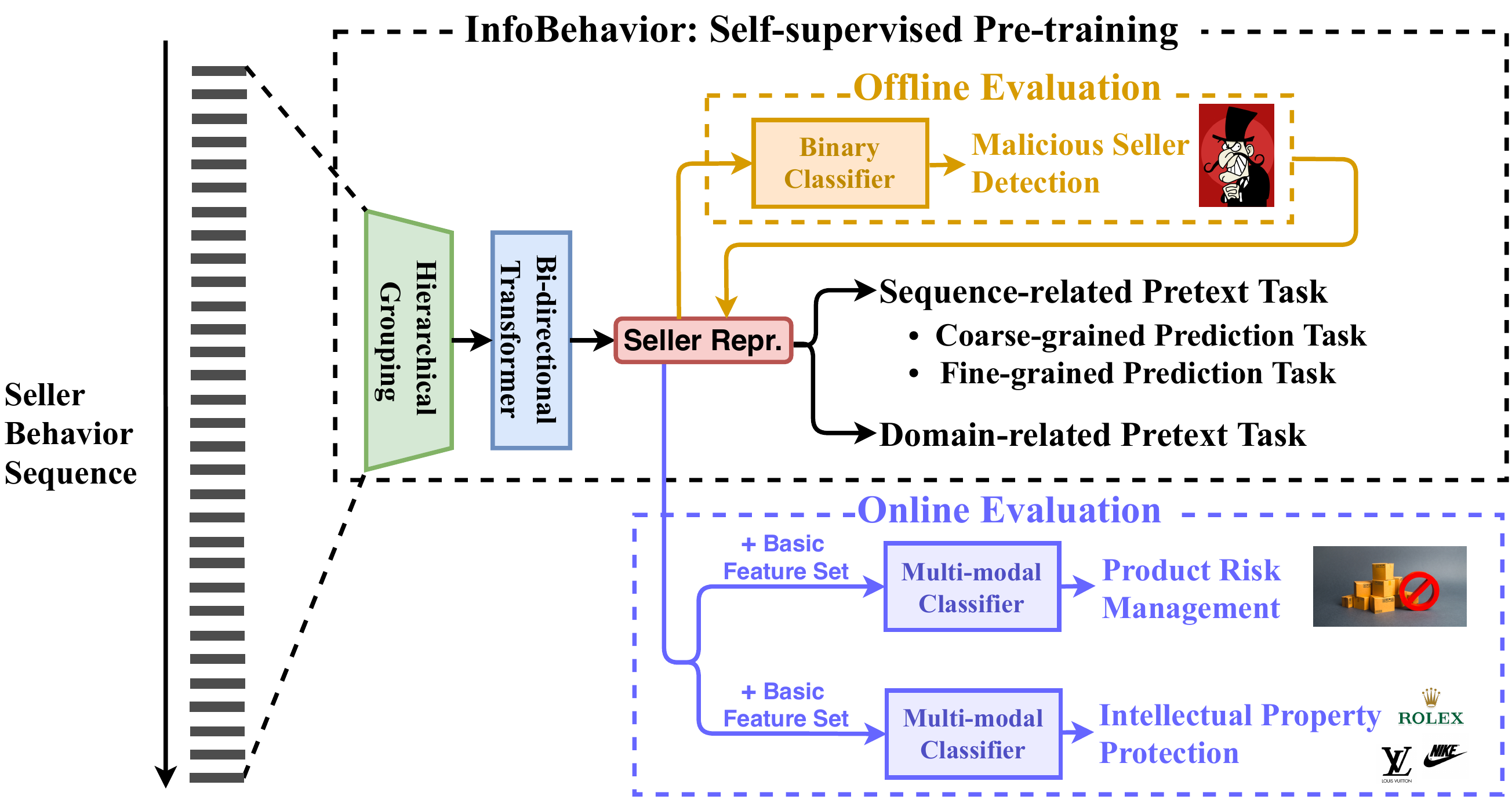}
		\caption{Overview of InfoBehavior model. InfoBehavior performs self-supervised training on two types of well-designed pretext tasks; Meanwhile, we leverage an offline task to evaluate its efficiency. The pre-trained InfoBehavior is directly applied to solve various online tasks.}
		\label{fig:overview}  
	\end{figure} 

    \textcolor{black}{Currently, self-supervised learning techniques have been developing rapidly due to the advance of the sophisticated deep neural network architectures and the powerful computing units. Various research domains yield distinct improvement, including Natural Language Processing \citep{devlin2018bert,yang2019xlnet,lan2019albert} and Computer Vision \citep{hjelm2018learning,oord2018representation}.} 
    Self-supervised learning techniques define pretext tasks that can be formulated by using only unlabeled data but do require higher-level semantic understanding in order to be solved~\cite{zhai2019s4l}.
    With the well-designed pretext tasks, it is proven that the generated representations possess remarkable generality. In terms of model selection, there are various network structures for sequence modeling, including CNN, RNN, and LSTM, etc. Compared with them, 
    Transformer model shows excellent capability in long-range dependency and computational efficiency. Naturally, it gradually becomes the first choice for self-supervised sequence modeling.

    In this paper, we introduce a self-supervised strategy InfoBehavior for representation learning of seller behavior sequence. \textcolor{black}{Our ultimate goal is to provide more robust behavior representations for the recognition of risky products.} Figure \ref{fig:overview} gives the overview of our InfoBehavior model.
    We first collect large-scale raw seller behavior sequences. Due to the existence of various external factors, seller behavior sequences may not strictly follow the ``rigid order sequence" assumption \citep{hu2017diversifying,wang2018attention}. Therefore, we decide to incorporate context information from both directions rather than merely left-to-right direction, and correspondingly select Bidirectional Transformer \citep{sun2019bert4rec} for seller behavior sequence modeling. Unlike conventional sequences, the length of a valid seller behavior sequence exceeds the affordability of the standard Transformer of commodity machines, \textcolor{black}{whose maximum length may reach hundreds of thousands}. 
    To solve this problem, we propose a hierarchical grouping strategy to deal with the ultra-long behavior sequence inputs. Hierarchical grouping consists of the minute-level aggregation and the CNN-based aggregation. 
    Minute-level aggregation is to aggregate a series of behaviors within the same minute into a meaningful aggregated behavior. The reason for minute-level time interval is the trade-off between information preservation and computation-storage cost. Subsequently, CNN-based aggregation is to integrate the representations of the aggregated behaviors within a fixed time span (such as several hours) into the high-level representation (span representations). After hierarchical grouping, the sequence length of span representations is in hundred level that is processable by commodity machines.
    
    We design two types of pretext tasks as in Figure \ref{fig:overview}. In terms of sequence-related pretext task, coarse-grained prediction task is to determine whether two month-level behavior sequences come from the same seller, and fine-grained prediction task is to correctly predict the masked behavior fragment given its context behaviors. Both of them are modeled via contrastive learning objectives. 
    Domain-related pretext task combines the expert prior knowledge, where we utilize the amount of seller's failure operation as the classification label. 
    After self-supervised pre-training, InfoBehavior model is able to generate reasonable behavior representations, and these representations can be directly fed into the detectors of downstream tasks. Our main contributions are three-fold:

    \begin{itemize}
    \item We are the first to consider self-supervised pre-training for seller behavior sequences and design three suitable self-supervision tasks for behavior representation learning;
    
    \item We propose a hierarchical grouping strategy for Transformer to model ultra-long sequence;
    
    \item Behavior representations generated by the pre-trained InfoBehavior effectively improve various info-security-related detection tasks at Alibaba.
    
    \end{itemize}

\section{RELATED WORK}
	\label{sec:related work}
    Due to the merits of large scale and rich information, behavior information has played a crucial role in the field of risk management. Conventional anomaly detection techniques \citep{huang2009visualization,suntinger2008event,li2017demalc} regard behavior sequence as spatiotemporal data, and emphasize the importance of location \citep{hao2006business} and timestamps \citep{olszewski2014fraud} in time series. The core concept of these works can be summarized as extracting anomalous patterns. \citet{huang2009visualization} leverage statistical methods to build customer profiles as suspected patterns from spatial, temporal and spectral information in historical databases. \citet{schaefer2011visual} employ clustering methods to gather events with similar temporal event patterns. An event pattern stands for an event sequence or event episode that displays interesting properties. Besides that, classification-based techniques \citep{olszewski2014fraud,argyriou2014fraud} and nearest neighbor-based techniques \citep{didimo2011advanced} are also widely used for anomaly detection task. These works highly depend on feature selection and expert knowledge. \textcolor{black}{Compared with them, we directly process the raw behavior sequences and automatically generate high-quality behavior representations for various downstream applications.}

	\begin{figure*}[t]
	  \centering
	  \includegraphics[width=0.85\linewidth]{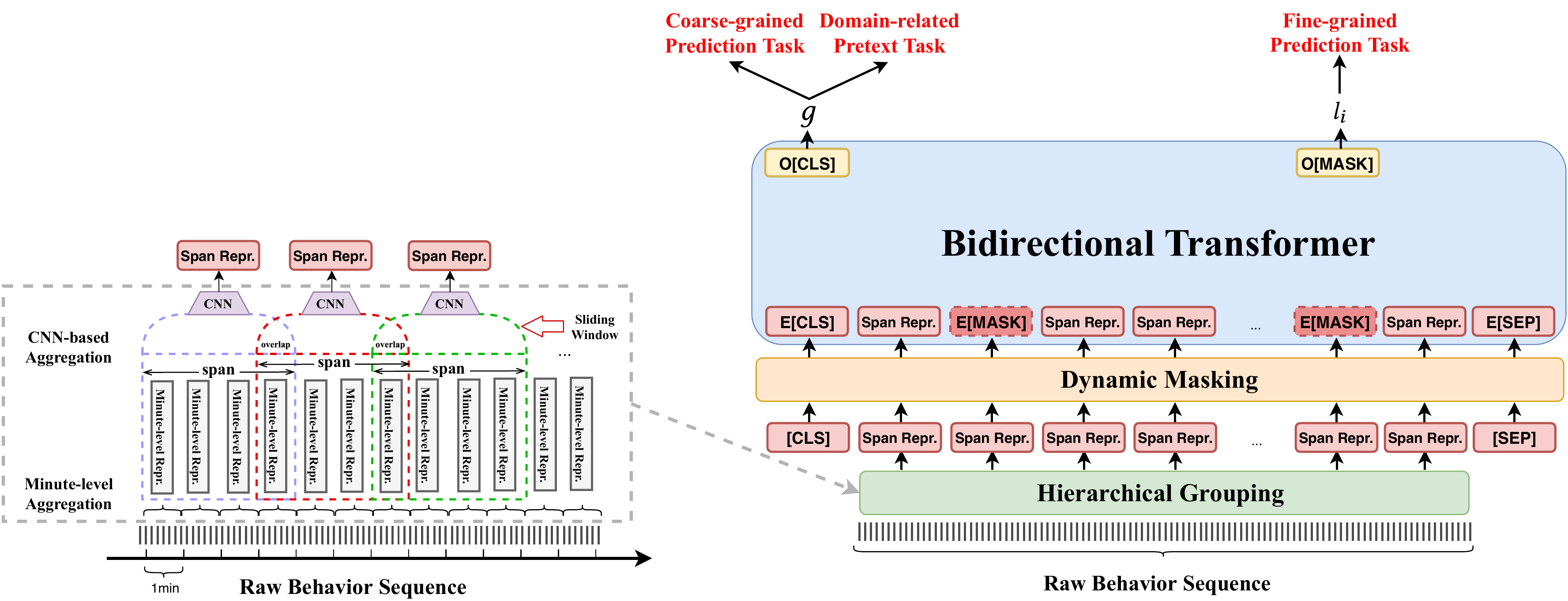}
	  \caption{The framework of InfoBehavior. Hierarchical grouping consists of two steps. Minute-level aggregation first calculates minute-level representations; CNN-based aggregation then generates high-level span representations via window sliding operation. Span representations are regarded as token-level representations and fed into Bidirectional Transformer. Three pretext tasks are used to learn behavior representations.}
	  \label{fig:architecture}  
	\end{figure*}
	
    \textcolor{black}{E-commerce companies pay increasing attention to the potential value involved in tremendous raw data.} Self-supervised learning has gradually become the key technique. In terms of behavior sequence modeling, previous works mainly focus on consumer behaviors for better recommendations, especially from click sequences and purchase sequences. \citet{hidasi2015session} adopt Gated Recurrent Units (GRU) to model click sequences for session-based recommendation; \citet{grbovic2018real} adopt the concept of contextual co-occurrence to learn low-dimensional representations of home listings and users in Airbnb. Recently, many works \citep{kang2018self,chen2019behavior} are inclined to use Transformer to capture the sequential signals because it is powerful in modeling long-distance dependency and building deeper network structure. Inspired by \emph{BERT} \citep{devlin2018bert}, \citet{sun2019bert4rec} propose a model, named BERT4Rec, to encode sequence information in a bidirectional way and achieve remarkable performance on next-item prediction task. \textcolor{black}{Different from that, we aim to model seller behaviors for info-security applications. Besides, compared with consumer behaviors, a valid seller behavior sequence is extremely long that greatly exceeds the modeling ability of current sequence networks.} Therefore, we propose a novel hierarchical grouping strategy to deal with ultra-long sequences.
    
    For sequence modeling, the conventional self-supervised objective \cite{devlin2018bert,sun2019contrastive} is to correctly predict the masked-out fragments of sequences. BERT4Rec \citep{sun2019bert4rec} defines a log-likelihood objective by softmax operation on product item vocabulary.
    However, in our case, the \textcolor{black}{number of high-level embeddings} generated by hierarchical grouping is infinite. Therefore, we adopt contrastive learning to design our training objective. InfoNCE~\cite{oord2018representation}, as a variant of Noise Contrastive Estimate~\citep{gutmann2010noise,mcallester2018formal,poole2019variational,ye2019unsupervised}, is commonly-used for self-supervised learning.
    \citet{trinh2019selfie} design an InfoNCE training objective to correctly select the correct patch among other ``distractor" patches sampled from the same image; \citet{bachman2019learning} also utilize InfoNCE objective to complete representation learning by maximizing mutual information between representations from different encoder layers. 
    The proposed InfoBehavior adopts InfoNCE training objective for both fine-grained prediction task and coarse-grained prediction task.

\section{METHODOLOGY}

    In this section, we introduce the proposed InfoBehavior in detail. The overview of the framework is presented in Figure~\ref{fig:architecture}. The following introduction is divided into four parts. First is the hierarchical grouping strategy that introduces how to generate the length-processable inputs from ultra-long behavior sequences. 
    Second is the sequence modeling part which briefly describes Bidirectional Transformer. The third part introduces three well-designed pretext tasks for encoding valuable behavior information into feature representations. The final part presents two training optimization strategies.

\subsection{Hierarchical Grouping Strategy}
    Longer behavior sequences can reflect more valuable information. In practical application, we usually extract features from seller behavior sequences of one month at least. \textcolor{black}{Assuming that a seller has one behavior in one minute, there will be 43,200 behaviors in 30 days. If this is a hot seller, hundreds of posting and editing behaviors can occur within one minute by using some auxiliary tools. The number of behaviors will exceed the million level in 30 days.} 
    Obviously, RNN-based models can not process this extremely long sequence due to gradient vanishing.
    As for Transformer model, the time and memory required grow quadratically with the increase of sequence length, which is also intractable for commodity machines. To solve this problem, we introduce a  hierarchical grouping strategy to convert raw seller behavior sequence into the \textcolor{black}{length-processable embedding sequence via two-step aggregation operation.}

\subsubsection{Minute-level Aggregation} 
\begin{figure}[t]
	  \centering
	  \includegraphics[width=0.8\linewidth]{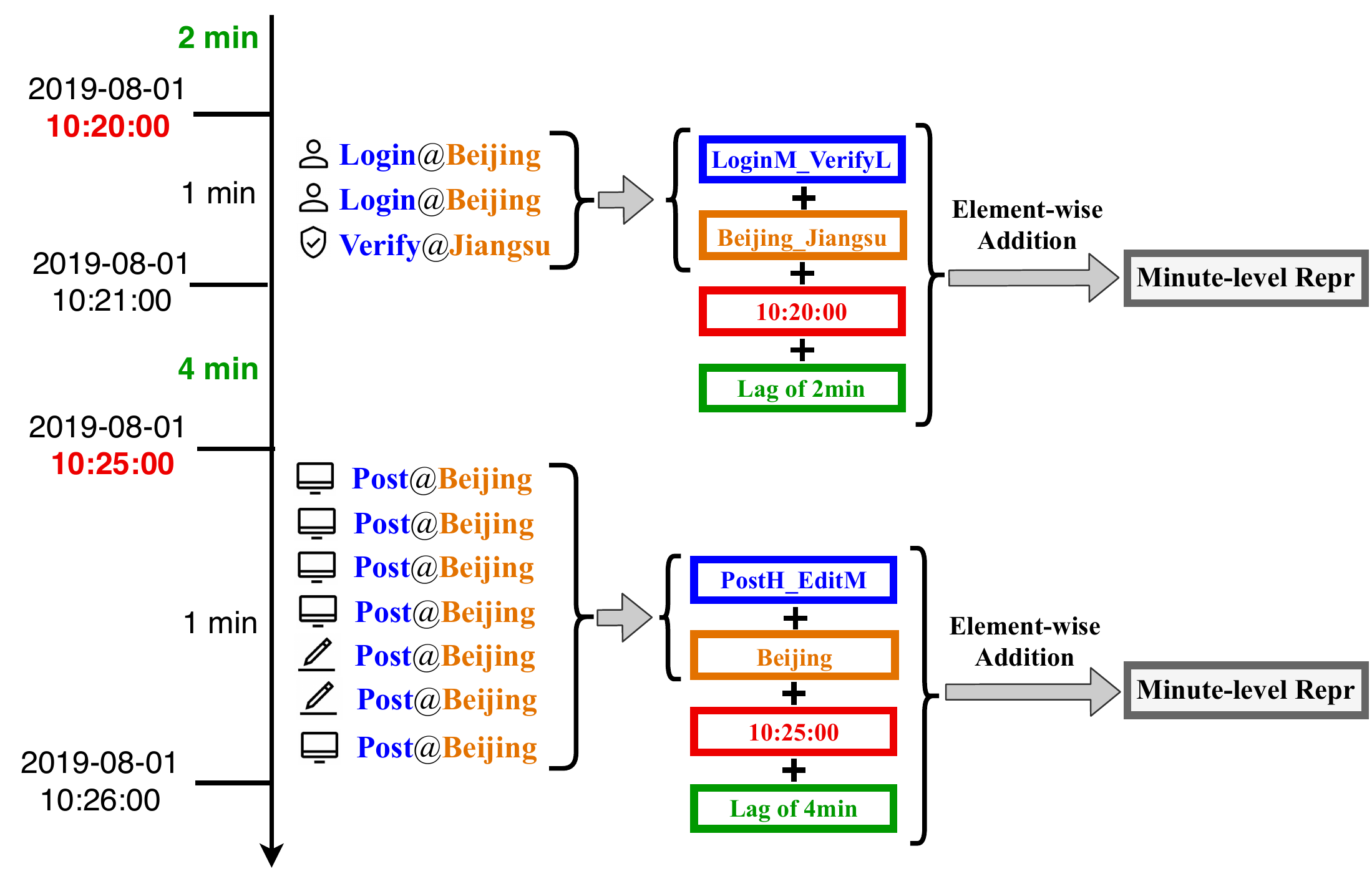}
	  \caption{Minute-level Aggregation. We aggregate four types of information in each minute, including behavior, geographical location, time point, and time lag. All this information in the same minute is encoded into a minute-level representation.}
	  \label{fig:MA}  
	\end{figure}

    We propose a simple minute-level aggregation to aggregate the raw behaviors within the same minute. Behavior information is usually considered as spatiotemporal data along with time and geographical information. For example, frequently switching login location is often regarded as high-risk behaviors. \textcolor{black}{The inputs} of minute-level aggregation consist of four different types of information as in Figure~\ref{fig:MA}, including behaviors, geographical location, the time point, and the time lag between two adjacent minute-level tokens. \textcolor{black}{The aggregation operation is exerted on behavior and geographical location.} 
    
    The phenomenon of repetitive behaviors is common, which causes the amount of minute-level behavior extremely large. Thus, we adopt a fuzzy definition. Based on statistical results, the number of the same operation within one minute is mapped to three levels: L (low), M (medium) and H (high). For behaviors like \emph{login}, \emph{verify}, \emph{modify} and\emph{ buy}, we maps 1 to L, 2 to M and values in [3,+inf) to H. For behaviors as \emph{post}, \emph{edit} and \emph{sell}, we maps 1 to L, values in [2,5) to M and values in [5,+inf) to H\footnote{Beside behavior types listed above, there are also several one-time behaviors like registration and application for opening a shop, which are unnecessary to adopt the fuzzy definition.}.
    As in Figure \ref{fig:MA}, if a seller \emph{Post} 5 products and \emph{Edit} 2 products in the same minute, the aggregated behavior is \emph{PostH\_EditM}.
    
    A seller may have multiple geographical locations in one minute due to multi-location management. Therefore, we adopt a similar aggregation rule but do not record the occurrence count, such as ``\emph{Beijing\_Jiansu}" and ``\emph{Beijing}" in Figure~\ref{fig:MA}. Besides, the time point and the time lag are directly mapped into tokens without the aggregation operation.
    
    \textcolor{black}{Four types of tokens mentioned above} are all vectorized and trained with the training process of InfoBehavior. \textcolor{black}{After minute-level aggregation, we get the minute-level representations,} and the maximum behavior sequence length is narrowed down to 1,440 in one day and 43,200 in 30 days.
	
\subsubsection{CNN-based Aggregation}    
The length of 43,200 is still unaffordable for Bidirectional Transformer, so we need to further aggregate the minute-level representations. We adopt a CNN-based aggregation operation. \textcolor{black}{As in Figure~\ref{fig:architecture}, given a sequence of minute-level representations, we first adopt the window sliding operation to obtain a series of behavior fragments\footnote{Noted that these splitting operations are exerted according to time series. Taking 30 days as an example, it can be understood that every minute-level representation sequence is first expended into the length of 43,200 by adding placeholders to non-behavior minute positions and then split.}, where each behavior fragment has the same interval of time span. We define each behavior fragment as a behavior span. To preserve the time continuity property of behaviors, we allow two adjacent spans remaining a certain overlap as in Figure \ref{fig:architecture}. 
    Then we utilize a convolutional neural network (CNN) model to generate span representation for each behavior span\footnote{To speed up the CNN-based aggregation, we first record the specific time information during minute-level aggregation; Then, the encoding operation of CNN is just exerted on the valid spans that have behaviors where the sequence order of minute-level representations is based on the recorded time information.}.
    The structure of this CNN consists of a multi-window convolutional layer and a max-pooling layer, which is derived from the TextCNN model \cite{kim2014convolutional}. With a rational setting of span size, the sequence length can be narrowed down to the length-processable scale for Transformer model. Therefore, span representations are directly used as the input representations for the following Bidirectional Transformer model.}

\subsection{Bidirectional Transformer}

    We adopt Bidirectional Transformer \cite{devlin2018bert} to model the sequence information. Compared with RNN-based models, Transformer has proved its superiority in modeling long-distance dependency. The structure of a Transformer layer consists of two sub-layers, multi-head self-attention layer, and position-wise feed-forward network. The core concept of the multi-head self-attention layer is the self-attention mechanism \cite{vaswani2017attention,xu2019multi}, which is beneficial to capture the dependencies between span representation pairs regardless of their distance in the sequences. Because self-attention is mainly based on linear operations, the function of the position-wise feed-forward network is to do non-linear transformation and generate high-level abstract representations. The capability of sequence modeling can be enhanced by easily stacking the Transformer layers. Besides that, the structure of Transformer layer is friendly for parallel computing which is critical for industrial applications on large-scale data training. 
    InfoBehavior aims to learn better representations of seller behavior sequence rather than predict the next behavior. Therefore, we utilize Bidirectional Transformer\footnote{Bidirectional Transformer is implemented by the attention mask trick. More specifically, the solution is to add a large negative value to the indices of masked tokens.},
    which is beneficial for modeling context information. As shown in Figure \ref{fig:architecture}, we add a special token [CLS] at the beginning of each input sequence. After the encoding of Bidirectional Transformer, the output representation of [CLS] is regarded as the \textbf{global} behavior representation $g$ of the whole seller behavior sequence. The output representations of [MASK] are regarded as the \textbf{local} behavior representations $l_i \in L$ corresponding to the input span representation $s_i$, where $L$ denotes the set of local behavior representations for one input behavior sequence.

\subsection{Pretext Tasks}

    \textcolor{black}{In this section, we introduce two sequence-related pretext tasks and a domain-related pretext task for seller behavior representation learning.} 
    The supervised signal of self-supervised tasks must be extracted from the data itself without human annotation.
    Due to the sequence structure, it is rational to predict the masked behaviors from its context behaviors~\cite{devlin2018bert,trinh2019selfie,bachman2019learning}. Therefore, our sequence-related pretext task is to select the correct masked behavior fragment against other ``distractor" behavior fragments. Two sequence-related pretext tasks, coarse-grained prediction task, and fine-grained prediction task are defined according to the granularity of the masked behavior fragments.
    Besides that, we design a domain-related task to encode the info-security prior knowledge.

\subsubsection{Coarse-grained Prediction Task}
\label{subsubsec:coarse} 
    Coarse-grained prediction task is designed based on the global behavior representation $g$ as in Figure~\ref{fig:loss_function}. We believe that seller behavior sequences can reflect the latent operation habit of sellers. More specifically, the behavior sequence pair derived from the same seller must be better matched than the pair from different sellers. 
    We leverage contrastive learning to define the training objective. As shown in Figure~\ref{fig:loss_function}, we adopt the structure of Siamese Network \citep{taigman2014deepface,zagoruyko2015learning,bertinetto2016fully}. Given a seller, we use $X$ to stand for a series of seller behaviors in the past long period of time. 
    Then, we define two data transformations $\mathcal{F}_1(\cdot)$ and $\mathcal{F}_2(\cdot)$. In our implementation, $\mathcal{F}_1(\cdot)$  and $\mathcal{F}_2(\cdot)$ perform the data extraction operation. $X_1 = \mathcal{F}_1(X)$ and $X_2 = \mathcal{F}_2(X)$ are the behavior sequences from the two adjacent months within $X$. Through the encoding of Bidirectional Transformer, we obtain two global behavior representations $g^1$ and $g^2$. 
    
    \citet{oord2018representation} have proven that it is feasible to complete feature representation learning by maximizing the mutual information between the representations that encode the underlying shared information, and they introduce a contrastive learning objective named InfoNCE. Analogously, $g^1$ and $g^2$ are from the same seller, so we utilize InfoNCE training objective to maximize the mutual information between $g^1$ and $g^2$, which is formulated as below,
    
    \begin{equation}
    \mathcal{L}_{coarse}(g^1,g^2,N_g^1) = -\log\frac{\exp(\Phi(g^1, g^2))}{\exp(\Phi(g^1, g^2))+ \sum_{j=1}^{|N_g^1|}\exp({\Phi(g^1, g_j^{\prime})})},
    \label{equ:coarse_part}
    \end{equation}
    
    \noindent where $\Phi(\cdot)$ denotes dot-product operation and $g_j^{\prime} \in N_g^1$ denotes the negative sample of $g_1$. We believe that different sellers have distinguishable behavioral habits, so the mutual information between positive pairs $<g^1, g^2>$ should be significantly larger than that between negative pairs $<g^1, g_j^{\prime}>$. 
    In our implementation, $g_j^{\prime}$ derives from the other sellers in the same mini batch~\citep{bachman2019learning}. Due to the asymmetry property of equation \ref{equ:coarse_part}, the overall definition of the coarse-grained prediction loss function is formulated as below,    
    \begin{equation}
    \mathcal{L}_{coarse} = \mathcal{L}_{coarse}(g^1,g^2,N_g^1) + \mathcal{L}_{coarse}(g^2,g^1,N_g^2)
    \label{equ:coarse}
    \end{equation}
    
    In summary, the coarse-grained prediction task is defined from the view of different sellers, which is to capture the key differences between different sellers’ behavioral habits.

\subsubsection{Fine-grained Prediction Task}
\label{subsubsec:fine}
\begin{figure}[t]
		\centering
		\includegraphics[width=0.76\linewidth]{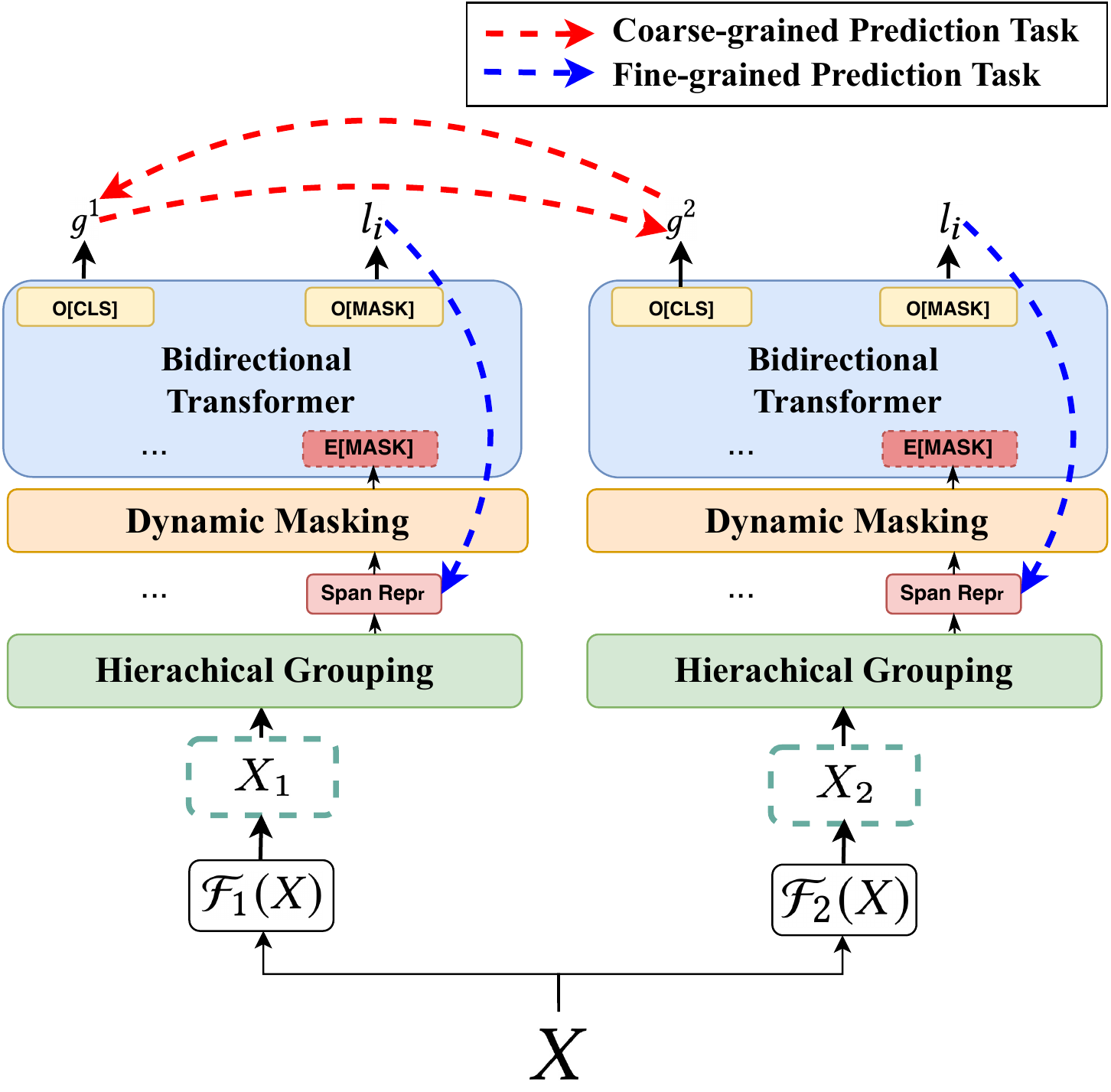}
		\caption{Contrastive Learning objectives of coarse-grained prediction task and fine-grained prediction task.}
		\label{fig:loss_function}  
	\end{figure}

    \textcolor{black}{Fine-grained prediction task is defined on the local behavior representation $l_i$. There exist latent logistic relationships within the behavior sequences, so it is possible to predict the masked behaviors via their context behaviors.}
    Based on that, we design a pretext task: predicting the masked span representation given its context span representations as Figure \ref{fig:loss_function}. Fine-grained prediction task is respectively exerted on $X_1$ and $X_2$. We also leverage contrastive learning to design the training objective. For each masked local behavior representation $l_i$, the positive representation is from its corresponding span representation $s_i$, and the negative representation $s_i^{\prime}$ is the masked span representation from the other sellers in the same mini batch. Accordingly, the loss function can be defined as below,

    \begin{equation}
    \mathcal{L}_{fine}(L, N_l) = -\sum_{i}\log\frac{\exp(\Phi(l_i, s_i))}{\exp(\Phi(l_i, s_i)+ \sum_{k=1}^{|N_l|}\exp{\Phi(l_i, s_k^{\prime})})},
    \label{equ:fine-part}
    \end{equation}
    
    \noindent where $s_k^{\prime} \in N_l$ and $N_l$ denotes the set of negative span representations. Considering $L_1$, $L_2$ from $X_1$, $X_2$, the overall training objective of fine-grained prediction is formulated as below,
    
    \begin{equation}
    \mathcal{L}_{fine} = \mathcal{L}_{fine}(L_1, N_l^1) + \mathcal{L}_{fine}(L_2, N_l^2)
    \label{equ:fine}
    \end{equation}
    
    In summary, the fine-grained prediction task is defined from the view of the same seller, which is to learn the logistic rationality of a series of behaviors ordered by time sequence.
	
\subsubsection{Domain-related Pretext Task}
In this section, we introduce a domain-related pretext task where labels are from the counting-based statistical results. It is inspired by previous feature engineering methods, and the goal is to encode the domain-specific prior knowledge into behavior representations. When dealing with seller behavioral data, the statistics-based features are commonly used. \textcolor{black}{Here we adopt the number of failure operations $c$ as the counting-based statistical indicators. These failure operations come from behaviors of \emph{login}, \emph{verify} and \emph{modify}. \textcolor{black}{For example, login failures are often caused by login environment changes. Often, the accounts of abnormal sellers are bought from black markets. Since these accounts are registered by others, abnormal sellers may not pass the verification every time and tend to modify the password or personal profiles.}
They can reflect anomalous behaviors to some extent. 
For example, abnormal sellers usually have many failed verification behaviors.} Label $c$ only can be inferred after processing the whole input behavior sequence, therefore we use global behavior representation $g$ as input. Besides that, we adopt a simple feed-forward network $G(\cdot)$ that consists of a linear layer and a softmax layer to calculate the classification probability $p_c \in P = {G(g)}$.  The training objective can be formulated as below: 
\begin{equation}
    \mathcal{L}_{count} = -\sum_{c=0}^{M-1} y_c \log(p_c)  %\big(y_k\log \hat{y_k} + (1-y_k)\log(1-\hat{y_k})\big),
\end{equation}
\noindent where $M$ is the cardinality of the label set and $y_c$ is a binary indicator (0 or 1). If class label $c$ is correct, $y_c=1$; Otherwise, $y_c=0$.

%In summary, domain-related pretext task is to encode the domain-specific information into representations via expert prior knowledge.

\subsubsection{Putting Them Together}
Combing these pretext tasks introduced above, the overall training objective of InfoBehavior can be formulated as below:

\begin{equation}
\mathcal{L} = \mathcal{L}_{coarse} + \mathcal{L}_{fine} + \mathcal{L}_{count}
\end{equation}

\subsection{Model Training}

In order to make the self-supervised training process more robust, we adopt two training optimization strategies: dynamic padding and dynamic masking. 
%Moreover, we present the computational analysis to demonstrate that InfoBehavior possesses obvious superiority of efficiency against previous solutions. 

\subsubsection{Dynamic Padding}
E-commerce companies need to process tremendous data each day. Therefore, time efficiency is regarded as a vital criterion of an algorithm. Based on this consideration, we adopt a dynamic padding strategy to reduce the computing effort of InfoBehavior pre-training. When modeling sequence data, the conventional way is to first count the maximum sequence length $d^{max}$ in the whole dataset, and then \textcolor{black}{pad each sequence to $d^{max}$.} 
Differently, our dynamic padding is to count the maximum sequence length $d^{batch}$ in each mini-batch. From our statistics, the behavior sequence length of most sellers is far less than $d^{max}$, because only a few hot sellers have the number of behaviors around $d^{max}$. Assuming that there are $n$ sellers, the saving computation effort of dynamic padding is proportional to $\sum_{i=1}^n { (d^{max} - d^{batch}_i)}$. It is more obvious when $n$ is on a larger scale.
We adopt a dynamic padding strategy both for the sequence modeling process of CNN-based aggregation and Bidirectional Transformer.

\subsubsection{Dynamic Masking}
\label{dynamic_masking}
Dynamic masking \cite{liu2019roberta} is a widely-used strategy for Transformer-based sequence modeling. Instead of assigning the masked positions during the data preparation stage \cite{devlin2018bert}, dynamic masking aims to dynamically change the masked positions during training. This strategy is beneficial for more training steps and larger scale of the training dataset.

%\subsubsection{Computational Analysis}
%\textcolor{green}{Coming soon......}

\section{EXPERIMENTS}

Analogous to the experimental setting of previous self-supervised representation learning methods \cite{devlin2018bert,bachman2019learning}, we first present the self-supervised training details of InfoBehavior model. Second, we design an \emph{abnormal seller detection} task as the offline evaluation task to evaluate the quality of behavior representations for the better configuration of InfoBehavior. In this part, we present the classification performance and a series of ablation studies. Finally, we introduce the online evaluation tasks. We use the real-time data from the e-commerce scenario, and validate the performance of pre-trained InfoBehavior for \emph{product risk management} task and \emph{Intellectual property preotection} task. 

\subsection{Self-Supervised Pre-training Stage}
\subsubsection{Dataset}
%$X_1 = \mathcal{F}_1(X)$ and $X_2 = \mathcal{F}_2(X)$
% [Sellers that need to be inferred for representations daily are those who are active in the past 30 days. Specifically, around 6 million record consisted of behavior sequences from past 30 days are being inferred everyday.]
    \textcolor{black}{The training dataset of the self-supervised stage consists of the behavior sequences of 2 consecutive months (August 2019 and September 2019) from 9,809,650 sellers, which is treated as the original behavior sequences $X$ corresponding to Figure~\ref{fig:loss_function}.
    %Then, we augment each $X$ into $X_1$, $X_2$ via $\mathcal{F}_1(\cdot)$, $\mathcal{F}_2(\cdot)$. 
    $\mathcal{F}_1(X)$ is to select the behavior sequence of August 2019 and $\mathcal{F}_2(X)$ is to select the behavior sequence of September 2019. 
    %Because $X_1$, $X_2$ belong to the same seller, we regard these two one-month behavior sequences as the augmented data pair. 
    Therefore, the input sequence of InfoBehavior is one-month behavior sequence.}

\subsubsection{Training Details}
\label{subsubsec:details}
    \textcolor{black}{InfoBehavior is implemented with PyTorch. We adopt AdamW~\cite{loshchilov2017decoupled} for parameter updates where weight decay is 0.01 and $\beta_1 = 0.9$, $\beta_2 = 0.999$. The learning rate is warmed up to a peak value of 1e-4 and then linearly decayed~\cite{liu2019roberta}. We set the warm-up proportion as 0.1.
    The gradient is clipped when the $l_2$ norm is larger than 1.0.} The processing of minute-level aggregation is implemented through DataWorks\footnote{https://www.alibabacloud.com/product/ide} by writing daily arranged SQL scripts. DataWorks is a Big Data platform product launched by Alibaba Cloud. 
    %It provides one-stop Big Data development, data permission management, offline job scheduling, and other features.
    \textcolor{black}{For the CNN-based aggregation, the span size is 360 (minute) and the stride is 340 (minute), with 20 minutes overlap between two adjacent spans. CNN encoder has the multi-window convolutional layer of 3, 4, 5 and the feature map dimension is 64. As for Bidirectional Transformer, we adopt a 4-layer Transformer with 4 attention heads. The dimensionality of each attention head is 48.} 
    Model training is performed on the PAI platform\footnote{https://www.alibabacloud.com/product/machine-learning}.
    \textcolor{black}{8 Nvidia 1080Ti GPU cards equipped with 50 CPU cores and 15GB memory for each card are used for model training.}
    The mini-batch size is 16 per GPU thus the overall batch size is 16$\times$8=128. The total number of training steps is 76,000, which costs 40 hours.
    %\textcolor{black}{Three self-supervised objectives including coarse-grained prediction task, fine-grained prediction task, and domain-related task are integrated through the training procedure.} 

\subsection{Offline Evaluation}

\subsubsection{Dataset}
The offline evaluation dataset is a binary classification dataset, which contains one-month behavior sequences of 502,355 sellers in October 2019. Among that, 2,361 sellers are the permanently-punished abnormal sellers, and the rest are the normal sellers that are randomly selected from the whole normal seller set. The dataset is then randomly split into training (80\%) and testing (20\%). During the training and evaluating stage, we regard abnormal sellers as positive samples and normal sellers as negative samples.

\subsubsection{Downstream Evaluation Setting}
During application, we first utilize the pre-trained InfoBehavior to calculate behavior representations. Then, these representations are used as inputs and fixed during the training of downstream classifiers. In terms of the network structure of classifiers, we use three different models: Logistics Regression (LR), GBDT and Multi-Layer Perceptron (MLP). Our \textbf{baseline} adopt the same structures of the classifier, but the inputs derive from the manually crafted features. 
These manual features derive from the same behavior sequences as the input of InfoBehavior.
%They are selected from the feature mart where features are used for online real-time models.

\subsubsection{Evaluation Metric}
We adopt two types of evaluation metrics for offline evaluation, including \textbf{NMI} (Normalized Mutual Information) and \textbf{AUC} (Area Under Curve). As suggested in \cite{jawahar2019does}, NMI \cite{strehl2002cluster} can be utilized to measure the quality of pre-trained representations without training classifiers.
% NMI \cite{strehl2002cluster,jawahar2019does} is to directly measure the quality of pre-trained representation based on the clustering results conditioned on class labels. 
Because the number of abnormal sellers is limited, we under-sample the normal sellers to maintain the positive-negative ratio around 1:20 to compute the NMI value. Then, we train three types of the classifier as mentioned above, and AUC is used to test the performance of classification. Since the training dataset is imbalanced, we over-sample~\cite{batista2004study} the positive samples by duplicating the positive data points for 4 times to balance the positive-negative sample ratio during the training procedure.

\subsubsection{Overall Performance}
\begin{table}
  \caption{Performance of Abnormal Seller Detection}
  \label{tab:MMD_classification}
  \begin{tabular}{ccc}
    \toprule
    \toprule
    Inputs & Classifier & AUC value\\
    \midrule
    Manual Feats.& LR & 0.8266 \\
    InfoBehavior Repr. & LR & 0.8993 \\
    Manual Feats. & GBDT & 0.9099 \\
    InfoBehavior Repr. & GBDT & 0.9146 \\
    InfoBehavior Repr. & MLP & \bf{0.9171} \\    
  \bottomrule
  \bottomrule
\end{tabular}
\end{table}

\begin{table*}[h!]
  \caption{Ablation Study of InfoBehavior Self-supervised Pre-training}
  \label{tab:ablation}
  \begin{tabular}{l|cccccc|cccc}
  	\toprule
  	\toprule
  	Ablation Part & \multicolumn{2}{c}{CNN-based Aggregation} && \multicolumn{3}{c}{Pretext Task} &  \multicolumn{1}{|c}{NMI} &\multicolumn{3}{c}{AUC} \\
  	\cmidrule{2-3}\cmidrule{5-7}\cmidrule{9-11}
  	  & Span Size & Stride & & Coarse. & Fine. & Domain-Related. &  & LR & GBDT & MLP\\
  	\midrule
  	Optimal Setting & 360 & 340 & & Yes & Yes & Yes & \bf{0.1825} & \bf{0.8993} & 0.9146 & 0.9171  \\
  	\midrule
  	\quad -Span Size & 180 & 160 & & Yes & Yes & Yes & 0.1394 & 0.8933 & 0.9185 & 0.8974  \\
  					 & 720 & 700 & & Yes & Yes & Yes & 0.1319 & 0.8934 & \bf{0.9188} & 0.8975  \\
  	\midrule
  	\quad -Span Overlap & 360 & 360 & & Yes & Yes & Yes & 0.1504 & 0.8845 & 0.9137 & 0.9142 \\
  	\midrule
  	\quad -Pretext Task & 360 & 340 & & No & Yes & Yes & 0.0492 & 0.8510 & 0.8937 & 0.8686  \\
  						   & 360 & 340 & & Yes & No & Yes & 0.1427 & 0.8925 & 0.9042 & 0.9064  \\
  						   & 360 & 340 & & Yes & Yes & No & 0.0851 & 0.8813 & 0.9121 & \bf{0.9180}  \\      	
    % bias1 & 360 & 340 & No & 128 & & Yes & Yes & 0.1705 & 0.8960 & 0.9276 \\  
    % bias2 & 360 & 340 & No & 64 & & Yes & Yes & 0.1761 & 0.8956 & 0.9158 \\  

  \bottomrule
  \bottomrule
\end{tabular}
\end{table*}

The comparison results of classification are shown in Table ~\ref{tab:MMD_classification}. First, it is obvious that the behavior representations from InfoBehavior outperform the manual features (Man. Feats.). It demonstrates that, with the assistance of three well-designed pretext tasks, InfoBehavior can encode valuable behavioral information into representations without human intervention. 
Moreover, manual features usually need to be specially designed for downstream tasks. Compared with that, behavior representations from pre-trained InfoBehavior can be directly used for various e-commerce info-security tasks, which effectively saves the application cost. Second, among these three types of classifiers, MLP achieves the best performance. This conclusion is consistent with previous works \cite{devlin2018bert}, i.e., even finetune the pretrained representation with very simple neural network, downstream tasks can yield great performance gains. 

\subsubsection{Ablation Study}
\textcolor{black}{In this section, we analyze the structure of CNN-based Aggregation and the impact of pretext tasks for self-supervised pre-training. The detailed ablation studies are presented in Table \ref{tab:ablation}. Our optimal setting is shown in the first row.}

%\begin{itemize}
%\item \textbf{Span Sizes} 
\stitle{Span Size} \textcolor{black}{When the span size is relatively short, CNN-based aggregation cannot encode enough information into the span representations; When the span size is relatively long, the information loss caused by convolutional encoding will increase. According to the comparison results (-Span Size) in Table~\ref{tab:ablation}, span size of 360 achieves the best performance (i.e., each span representation encodes the behavior fragment of 360 minutes). }
%, which means the behavior fragment of 360 minutes (6 hours).} 

%\item \textbf{Span Overlap}
\stitle{Span Overlap} \textcolor{black}{The comparison results (-Span Overlap) demonstrate that it is necessary to remain span overlap between adjacent spans. When the span size equals to stride, it means that spans are non-overlapping and the performance is quite worse than the overlapping version. Under the same span size, more span overlap will increase the computational cost. As in Table~\ref{tab:ablation}, considering the trade-off between performance and computational cost, we set the stride as 340 when span size equals to 360.} 
%Because hierarchical grouping splits seller behavior sequence based on time interval, contiguous behaviors are split into different spans if there is no overlap between adjacent spans.  
%It must influence the sequence coherence to some extents. Therefore, we adopt the span overlap to preserve the relationship between adjacent spans.}

%\item \textbf{Pretext Tasks} 
\stitle{Pretext Tasks} \textcolor{black}{Table~\ref{tab:ablation} (-Pretext Task) compares the influence of three proposed pretext tasks for the quality of behavior representations. First, we find that the coarse-grained prediction task plays the most important role in the self-supervised pre-training of InfoBehavior. It helps InfoBehavior to learn the global view of sellers' operation habits, and implicitly indicates the property of their posted products. Moreover, the global behavior representation $g$ is directly used for downstream tasks. 
Second, we find that the fine-grained prediction task brings the least performance gain. This phenomenon is inconsistent with the language self-supervised model (i.e., BERT), and we found that the contextual dependency of behavior sequences is relatively weaker than language sequences.
%provides supplement knowledge for representation learning. It focuses on modeling the logistic coherence within behavior sequences. 
Third, for evaluating the rationality of domain-related pretext task, we build a test dataset where behavior sequences do not exist in the training data, and it yields the classification accuracy (i.e., correctly predict the number of failed operations in a behavior sequence) of 97.52\% with the pre-trained InfoBehavior model. 
%It demonstrates the rationality of this pretext task. 
Table~\ref{tab:ablation} demonstrates that our domain-related pretext task brings the improvement of 0.1 on the value of NMI. On the whole, three well-designed pretext tasks effectively enhance the performance of InfoBehavior.}

%\end{itemize}

\subsection{Online Evaluation}
\begin{figure}[t]
    \centering
    \includegraphics[width=0.8\linewidth]{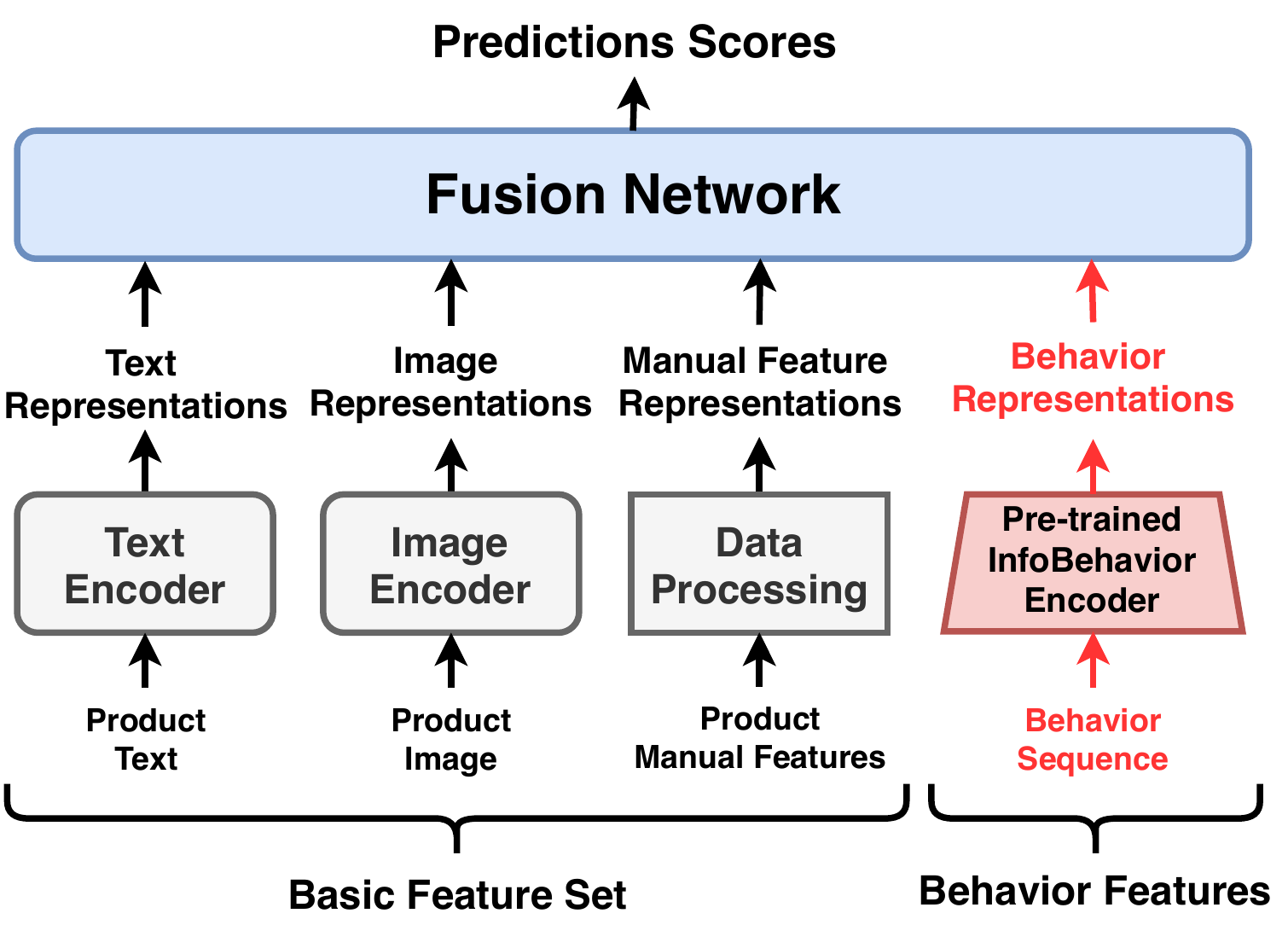}
    \caption{The framework of online system}
    \label{fig:online_evaluation}  
\end{figure}

In this section, we focus on evaluating InfoBehavior method on two real online tasks: Product Risk Management and Intellectual Property Protection. 
%Behavior information is vital in the info-security area. Abnormal sellers usually have abnormal behavioral patterns. For example, some abnormal sellers buy accounts from black markets; Therefore, their accounts would have login behaviors by different people at different time periods. In previous applications, behavior information is usually reflected in the way of the statistical feature. However, it has two drawbacks in deriving the statistical feature. First, the length of the behavior sequence is often very long, which means that it is hard for technicians to analyze and the cost of human annotation is very expensive. Second, behavior information belongs to spatiotemporal data. If temporal and spatial information are both considered with behavior information, the feature space will be extremely large. InfoBehavior effectively makes up these deficiencies.
%\noindent \textbf{Downstream Task Inference}\quad 
In the practical application, behavior representations are re-computed using the pre-trained InfoBehavior model daily and then saved into a partitioned table. Figure~\ref{fig:online_evaluation} presents the application of behavior representations on our online systems. \textbf{Basic feature set} includes text representations, image representations, and various domain-related manual features. In terms of behavior features, we apply two settings: \textbf{manual behavior features} and \textbf{InfoBehavior representations}. For a fair comparison, two different behavior features are generated from the seller behavior sequences of the same period.

\subsubsection{Performance of Product Risk Management}
The product risk management system aims to automatically detect and remove online restricted products. 
%For example, products that are related to  ``Accounts of Other Websites or Applications" or ``Gambling Auxiliaries" are strictly prohibited to be sold. 
%In this scenario, basic feature set mainly contains two parts of feature: textual features from product titles and detailed descriptions and visual features from product images. Manual Behavior Features involve the statistical features from seller behaviors. 
In our e-commerce scenario, products related to  ``Accounts of Other Websites or Applications" and ``Gambling Auxiliaries" are strictly prohibited to be sold and suffer from serious problems of content-based confrontation.
Therefore, we present the performance on these two tasks. We train the online fusion network and compare the performance with two different baseline settings as in Table~\ref{tab:PRM}. 
To satisfy business requirements, the online product risk management system needs high precision to avoid punishing normal sellers. Therefore, we are required to evaluate the performance of recall under the precision of 95\% that is denoted as \emph{Recall@95\%Precision} (R@95\%P in Table~\ref{tab:PRM}). Besides, we also leverage \emph{F1} as an evaluation criterion.
Compared with the basic feature set, InfoBehavior behavior representations brings significant performance gains for both tasks on both two evaluation criteria.
Particularly, InfoBehavior representations show notable superiority against the manual behavior features, which respectively achieves the recall improvement of 1.84\% and 3.86\% for Accounts and Gamble. It demonstrates that InfoBehavior is a better solution for behavior representation learning.

\subsubsection{Performance of Intellectual Property Protection}
E-commerce companies also face the problems of counterfeit products. To protect the right of normal sellers, like ``NIKE”, ``Rolex”, ``Louis Vuitton”, ``Samsonite”, ``Dior” and ``CHANEL”, e-commerce companies urgently need to build an effective Intellectual Property Protection system. 
Clothing \& Accessories, Luggage, and Cosmetics are three product categories that have serious problems of counterfeit products. We utilize InfoBehavior behavior representations for Intellectual Property Protection in these categories \textcolor{black}{aiming to detect sellers who tend to sell counterfeit products}, and evaluate the performance compared with previous settings.
\textcolor{black}{Therefore, the seller-related manual features play more important roles in this basic feature set.}
The comparison results are shown in Table \ref{tab:CPI}. \textcolor{black}{Since abnormal sellers detected by the system will be double-checked by auditors, the precision of 50\% is acceptable for this scenario.} Here we use \emph{Recall@50\%Precision}  (R@50\%P) and \emph{F1} as evaluation criteria. 
The results of manual behavior features indicate that behavior information is critical for this scenario. Compared with that, InfoBehavior representations bring further impressive improvement. With the assistance of InfoBehavior behavior representations, the Intellectual Property Protection system achieves recall improvement of 4\% - 5\% than the basic feature set. It proves that InfoBehavior representations encode more reasonable behavioral information than manual behavior features and greatly benefit the Intellectual Property Protection system. 

\begin{table}
	\caption{Model Comparison for Product Risk Management}
	\label{tab:PRM}
	\begin{tabular}{llcc}
		%\toprule
		\toprule
		Category & Features & R@95\%P  & F1 \\
		\midrule
		\multirow{2}{*}{Accounts}  & Basic Feature Set & 52.10\% & 0.6729  \\
		& \,\,\,+Manual Behavior Feats. & 53.78\% & 0.6868 \\
		& \,\,\,+InfoBehavior Repr. & 55.62\% & 0.7016 \\
		% \midrule
		% \multirow{2}{*}{Drugs}  & Man. Feats. & 25\% & 0.3333 & 0.8557 \\
		% & Man. Feats. \& Emb. & 30\% & 0.3750 & 0.8631 \\
		\midrule
		\multirow{2}{*}{Gamble}  & Basic Feature Set & 79.92\% & 0.8681  \\
		& \,\,\,+Manual Behavior Feats. & 80.12\% & 0.8693  \\
		& \,\,\,+InfoBehavior Repr. & 83.98\% & 0.8915  \\    
		
		\bottomrule
		%\bottomrule
	\end{tabular}
\end{table}

\begin{table}
	\caption{Model Comparison for Intellectual Property Protection}
	\label{tab:CPI}
	\begin{tabular}{llccc}
		%\toprule
		\toprule
		Category & Features & R@50\%P  & F1\\
		\midrule
		\multirow{2}{*}{Cloth. \& Acc.}  
		& Basic Feature Set & 36.97\% & 0.4251  \\
		& \,\,\,+Manual Behavior Feats. & 37.65\% & 0.4295 \\
		& \,\,\,+InfoBehavior Repr. & 41.48\% & 0.4534  \\
		\midrule
		\multirow{2}{*}{Luggage} 
		& Basic Feature Set & 26.38\% & 0.3454  \\
		& \,\,\,+Manual Behavior Feats. & 29.54\% & 0.3714 \\
		& \,\,\,+InfoBehavior Repr. & 30.91\% & 0.3820  \\
		\midrule
		\multirow{2}{*}{Cosmetics}
		& Basic Feature Set & 46.32\% & 0.4809   \\
		& \,\,\,+Manual Behavior Feats. & 48.39\% & 0.4918 \\
		& \,\,\,+InfoBehavior Repr. & 51.96\% & 0.5096  \\
		
		\bottomrule
		%\bottomrule
	\end{tabular}
\end{table}

\begin{spacing}{0.2}
	\subsection{Discussion}
\end{spacing}
\subsubsection{How to Learn Better Behavior Representations}
~\\
\stitle{Add more behavior types} The current version of InfoBehavior only considers some basic behavior types, such as \emph{register}, \emph{open online shop}, \emph{login}, \emph{verify}, \emph{modify}, \emph{edit product}, \emph{post product}, etc. They are main behavior types of seller and relatively easy to incorporate. Besides these, there also exist some behavior types that can implicitly provide valuable information, for example, \emph{decorate shop}, \emph{edit profile}, \emph{review product} and \emph{chat on Alitalk}. More varied behavior information is beneficial for more accurate depiction of sellers' habits, but in the meanwhile, it increases the difficulty of representation learning.

\stitle{Consider longer behavior sequences}
Both self-supervised training and downstream prediction are based on one-month behavior sequences. Even though it is a relatively long time period, some favorable factors still cannot be considered, such as seasonal variation and festival-oriented promotional activities. Obviously, longer behavior sequences can provide more information and reflect more generalized and accurate behavioral habits of sellers, which is beneficial for better behavior representations. But at the same time, it increases the computation and storage costs.

\stitle{Design more domain-related pretext tasks}
The ablation study from Table~\ref{tab:ablation} demonstrates that domain-related pretext task further improves the performance. It proves that domain-related prior knowledge is conducive to better self-supervised pre-training. Besides the number of failed operations, there are many other possibilities to build domain-related pretext task, such as the number of IP-address switch and the change of the affiliated ID card number.
%In future, we will consider 
%Based on that, if we can design more reasonable domain-related pretext tasks, it will definitely encode more valuable information into behavior representations.

\subsubsection{How to Better Apply Behavior Representations}
~\\
In the \textcolor{black}{info-security scenario}, current models leverage various types of input features, including textual representations, visual representations and plenty of hand-crafted features. However, model training with these features exists the over-fitting problem (i.e., overfit to textual, visual and other hand-crafted features, but not the behavior representations derived by InfoBehavior), which seriously impacts the generalization ability of the model. Under this condition, it cannot release the full potential of behavior representations. 
%To make better use of behavior representation, we provide a solution. 
%During downstream application, we can add a dynamic masking layer before the fusion network. 
To make better use of behavior representation, we suggest adding a dynamic masking layer before the fusion network in the training of downstream tasks.
Different from the setting in Section~\ref{dynamic_masking}, this dynamic masking layer randomly masks several feature inputs with different weights, and the probability for masking InfoBehavior representations is relatively small.
% but always retains behavior representations. 
The strategy can force the model to extract information from behavior representations to the most extent. We believe that behavioral information can effectively boost model generalization.

\section{Conclusion}
In this paper, we propose a self-supervised representation learning method, InfoBehavior, to generate meaningful behavior representations. Because seller behavior sequences are usually extremely long, we introduce a hierarchical grouping strategy to convert raw behavior sequences into length-processable span representations. Moreover, we design two types of pretext tasks. The sequence-related pretext tasks learn the behavioral habits and the logistic coherence of behavior sequence; The domain-related pretext task is to encode the domain-specific prior knowledge.
To the best of our knowledge, we are the first to consider self-supervised learning on seller behaviors and provide the first viable solution for Transformer to deal with ultra-long behavior sequences.
Experimental results demonstrate that behavior representations generated by the pre-trained InfoBehavior can be directly used or integrated with the representations from other side information to support a wide range of downstream tasks. Finally, the InfoBehavior has been deployed online to protect millions of consumer's rights every day.

%\begin{acks}
%We are appreciated for the assistance of Leishi Xu, Wei Peng, Jintian Deng, Bin Liu, who are from model-deploying platform team. With their help, real-time applications of InfoBehavior are implemented with great efficiency and efficacy.
%\end{acks}

\bibliographystyle{ACM-Reference-Format}
\bibliography{InfoBehavior_Reference}

\appendix

\end{document}